\documentclass{article}
\usepackage[preprint]{neurips_2026}

\usepackage[utf8]{inputenc}
\usepackage[T1]{fontenc}

\usepackage{amsmath}
\usepackage{amssymb}
\usepackage{amsfonts}
\usepackage{mathtools}

\usepackage{graphicx}
\usepackage{booktabs}
\usepackage{multirow}
\usepackage{makecell}
\usepackage{subcaption}
\usepackage{wrapfig}

\usepackage{algorithm}
\usepackage{algorithmic}
\usepackage{float}

\usepackage{nicefrac}
\usepackage{microtype}
\usepackage{xcolor}
\usepackage{url}
\usepackage{hyperref}

\graphicspath{{figures/}}

\newcommand{\method}{GeoTopoDiff}

\newcommand{\todofig}[2]{%
  \IfFileExists{figures/#1}{\includegraphics[width=#2]{#1}}{%
    \fbox{\begin{minipage}[c][0.26\textheight][c]{0.94\linewidth}
    \centering\small Figure file \texttt{figures/\detokenize{#1}} not provided.\\
    Replace this placeholder with the corresponding figure.
    \end{minipage}}}}

\newcommand{\todosmallfig}[2]{%
  \IfFileExists{figures/#1}{\includegraphics[width=#2]{#1}}{%
    \fbox{\begin{minipage}[c][0.18\textheight][c]{0.94\linewidth}
    \centering\small Figure file \texttt{figures/\detokenize{#1}} not provided.\\
    Replace this placeholder with the corresponding figure.
    \end{minipage}}}}

\title{
GeoTopoDiff: Learning Geometry--Topology Graph Priors through Boundary-Constrained Mixed Diffusion for Sparse-Slice 3D Porous Reconstruction
}

\author{
Yue Shi\\
Manchester Metropolitan University, Manchester M1 5GD, U.K.\\
\texttt{y.shi@mmu.ac.uk} \\
\And
Peng Wang \\
University of Surrey, Surrey, GU2 7XH, U.K. \\
\texttt{peng.wang@surrey.ac.uk} \\
\And
Mingzhe Yu \\
Johnson Matthey
Billingham, TS23 1LB, U.K. \\
\And
Yunlong Zhao \\
Imperial College London
London SW7 2AZ, U.K. \\
\And
Li Liu \\
Johnson Matthey
Billingham, TS23 1LB, U.K. \\
\And
Gareth D Hatton \\
Johnson Matthey
Billingham, TS23 1LB, U.K. \\
\And
Yan Lyu \\
University of Surrey, Surrey, GU2 7XH, U.K. 
 \\
\And
Liangxiu Han\\
Manchester Metropolitan University, Manchester M1 5GD, U.K.\\
}

\begin{document}

\maketitle

% ...

\begin{abstract}
Diffusion-based voxel prior modelling is challenging for the reconstruction of large-scale 3D porous microstructures.
Due to the demanding requirements for simultaneously modelling both the continuous pore morphology and the discrete pore-throat topology, the diffusion models require fully observed $\mu$CT scans to provide topology-faithful priors, which results in an inherent trade-off among throughput, topological fidelity, and field of view in practical industrial applications.
We propose GeoTopoDiff, a graph diffusion-based framework for reconstructing 3D porous microstructures from sparse $\mu$CT slices.
GeoTopoDiff transfers the learning of diffusion priors from a voxel-based space to a mixed graph state space, which simultaneously encompasses continuous pore geometry and discrete pore-throat topology.
A topology-aware partial graph prior from sparsely observed $\mu$CT slices is introduced to constrain the reverse denoising process. 
Experiments on anisotropic PTFE and Fontainebleau sandstone show that GeoTopoDiff reduces morphology-related errors by \(19.8\%\) and topology-sensitive transport errors by \(36.5\%\) on average.
Our findings suggest that the mixed graph state space promotes the diffusion denoising process to reduce posterior uncertainty under a sparse observations.
All models and code have been made publicly available to facilitate the exploration of diffusion models in the field of 3D porous microstructures simulation.

\end{abstract}

\section{Introduction}

Diffusion models provide a powerful class of learned priors for ill-posed inverse problems. 
By learning a time-dependent denoising score of an underlying data distribution, they enable reverse sampling from noise to data and can be combined with an observation model to solve conditional reconstruction tasks \citep{Ho2020DDPM}. 
Given a partial or degraded observation \(y=\mathcal{A}_{\Omega}(x)+\epsilon\), where \(\mathcal{A}_{\Omega}\) denotes an acquisition operator and \(\epsilon\) denotes measurement noise, diffusion inverse solvers use the learned prior \(p_{\theta}(x)\) together with the likelihood \(p(y\mid x)\) to sample plausible reconstructions from the posterior \(p(x\mid y)\). 
This posterior-sampling view is especially useful when the inverse problem is underconstrained, and multiple reconstructions may be consistent with the same observation. 
It has motivated diffusion-based methods for a wide range of image inverse problems, including deblurring, super-resolution, magnetic resonance imaging, and computed tomography reconstruction.

Micro-computed tomography (\(\mu\)CT) provides non-destructive, high-resolution imaging of microstructures and has become an important tool for materials characterization, industrial digital twin, and geoscience modeling \citep{song2024diffusionblend,Cnudde2013HighResolution,Wildenschild2013Xray}.
In pore-scale simulation, \(\mu\)CT scans provide important 2D insight into material characterizations of porosity, connectivity, and permeability \citep{Blunt2013PoreScale}. 
Diffusion model provides a promising tool to learn a 3D porous interior prior from 2D \(\mu\)CT scans \citep{Hui2024MicroDiffusion, gao2022cocodiff}. 
However, to learn material-specific priors, conventional voxel-space diffusion reconstruction pipelines rely on high-quality, fully observed \(\mu\)CT scans  \citep{Lee2024Micro3Diff}.
The fully observed \(\mu\)CT scans lead to a trade-off among scan throughput, field of view, and reconstruction fidelity \citep{Zhang2021SliceToVoxel}, which constrains the deployment of generative reconstruction in real-world industrial tasks.
In contrast, conditional diffusion pipelines based on sparse observations of \(\mu\)CT boundaries (e.g., top and bottom \(\mu\)CT slices) can provide more efficient inference and are better suited for high-throughput industrial imaging \citep{xia2022lowdose, gao2022cocodiff}.
Yet, sparse boundary conditions further complicate the ill-posed nature of the diffusion inverse problem; thus, the same boundary observations can lead to various visually plausible reconstructions that possess contradictory pore-throat graph topologies. 
The key limitation is latent space representation. 
Existing approaches perform reverse denoising in voxel occupancy space, which captures local morphology but fails to explicitly model the discrete pore-throat topology. 

\begin{figure}[t]
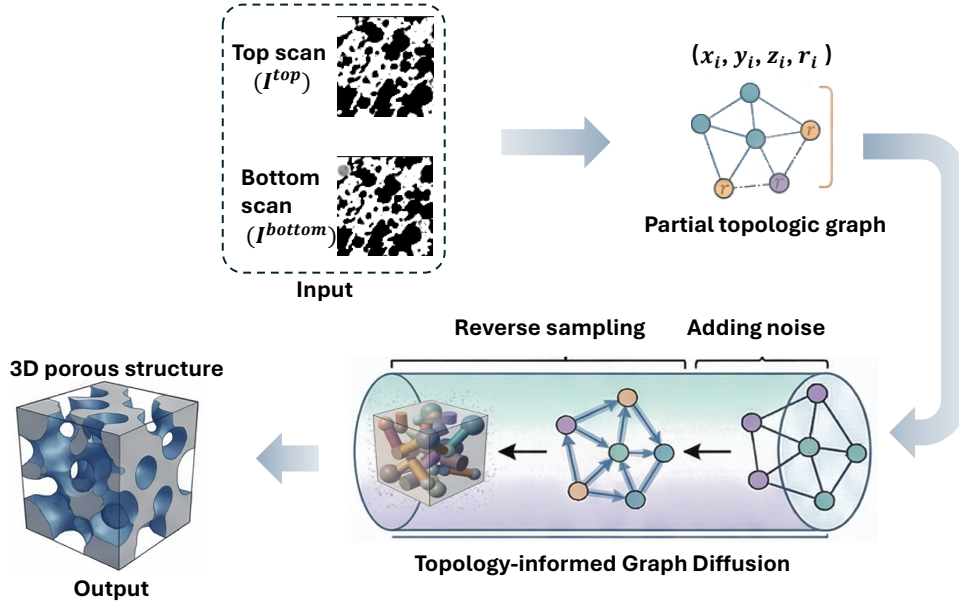

    \centering
    \todofig{1_overview.pdf}{0.95\linewidth}
    \caption{
    Overview of \method{}. 
    Sparsely observed top and bottom \(\mu\)CT slices are segmented and encoded as a partial topological boundary graph. 
    \method{} performs topology-aware graph diffusion in a mixed state space of continuous pore geometry and discrete pore-throat topology, and decodes the denoised graph into a 3D porous structure while enforcing boundary consistency.
    }
    \label{fig:overview}
\end{figure}

To address this limitation, we propose a graph diffusion framework, named \method{}, for topology‑faithful reconstruction of 3D porous microstructures based on sparse top and bottom \(\mu\)CT slices.
Instead of learning in voxel occupancy space, \method{} learns continuous pore geometry and discrete pore–throat topology in a mixed graph state space.
A topology-aware partial boundary graph extracted from top and bottom \(\mu\)CT slices is introduced to constrain the posterior of reverse denoising over the unobserved interior. 
By incorporating this partial graph prior into the image denoising process, \method{} reduces posterior uncertainty in unobserved internal regions and facilitates topologically consistent reconstruction under sparse observation conditions.
An overview of GeoTopoDiff is illustrated in Fig.~\ref{fig:overview}.

Our contributions are as follows.
\begin{itemize}
    \item We formulate sparse-boundary \(\mu\)CT slice-based 3D porous interior reconstruction as a mixed continuous-discrete diffusion inverse problem, which preserves both continuous pore geometry and discrete pore-throat topology.

    \item We propose \method{}, which transfers diffusion prior learning into a mixed graph state space and employs a topology‑aware partial graph prior to constrain the reverse denoising process for unobserved interior regions.

    \item We evaluate \method{} on Fontainebleau sandstone and anisotropic PTFE, showing improved morphology, graph topology fidelity, and transport-oriented consistency over representative baselines.
\end{itemize}
\section{Related Work}

\paragraph{Diffusion priors and 2D-to-3D inverse solvers.}
Denoising diffusion models learn generative priors by reversing a noise corruption process, and have become powerful tools for posterior sampling in inverse problems \citep{Ho2020DDPM, Chung2023DPS}. 
For degraded observations, diffusion inverse solvers integrate learned image priors with observed boundary conditions to enable probabilistic reconstruction.
However, directly learning and sampling complete 3D diffusion prior models is computationally intensive, as the memory and sampling costs are significantly higher than those of 2D images.
A common strategy is therefore to use 2D priors for 3D reconstruction. 
DiffusionMBIR trains a diffusion model on axial 2D slices and combines posterior sampling with total variation regularisation along the remaining directions to promote 3D reconstruction consistency \citep{Chung2023DiffusionMBIR}. 
DDS combines diffusion sampling with Krylov subspace techniques and data consistency constraints, significantly reducing the number of neural network function evaluations (NFEs) needed for CT reconstruction tasks \citep{Chung2024DDS}. 
TPDM further improves 3D coherence by using pre-trained perpendicular 2D diffusion models to reconstruct 3D medical volumes from multiple orthogonal planes \citep{Lee2023TPDM}. 
More recently, \citep{song2024diffusionblend} developed DiffusionBlend to learn 2D CT slice patch priors and fuse location-aware patch scores to generate full-volume scores for sparse-view and limited-angle 3D reconstruction. 
These methods demonstrate that diffusion priors based on 2D slices can render 3D inverse problems solvable.
Nevertheless, their prior distributions remain primarily confined to voxel-based latent space designs.
For porous media, since the fidelity of 3D reconstruction depends on the plausibility of pore-throat connectivity, it is challenging to leverage 2D slice-based prior distributions to explicitly parameterise the discrete topological structure of long-range connected pathways.

\paragraph{\(\mu\)CT-based 3D reconstruction and porous microstructure generation.}

Micro-computed tomography (\(\mu\)CT) provides non-destructive, high-resolution 3D imaging of internal microstructures and is widely used in materials characterization, industrial inspection, and geoscience. \citep{Withers2021XrayCT, Cnudde2013HighResolution}. 
It enables the digital reconstruction of invisible internal structures to support subsequent segmentation, defect analysis, and transport simulation. 
However, high-resolution full-volume \(\mu\)CT acquisition remains costly. Increasing spatial resolution typically leads to a reduced field of view (FoV), extended scanning times, and higher expenses for reconstruction and analysis
These limitations have prompted the development of deep learning-based reconstruction models, enabling them to infer three-dimensional internal structures from limited observational data.

Porous media represent a particularly challenging domain in \(\mu\)CT-based reconstruction, as their functional properties are governed by the pore network structure (e.g. porosity, connectivity, permeability, etc) \citep{Blunt2013PoreScale, Torquato2002RHM}. 
While classic approaches, such as descriptor matching and multi-point statistics, offer interpretable reconstruction priors, they are generally computationally costly and face significant challenges in handling long-range connectivity under sparse observations. \citep{Yeong1998RandomMediaII, Bostanabad2018Review}. 
Deep generative methods, including GANs, VAE-based models, and diffusion models, have improved 3D microstructure synthesis from 2D observations \citep{Chi2023CVAEGAN,Fu2023SINN,Lyu2024MicrostructureDDPM,Lee2024Micro3Diff}. 
However, these methods primarily learn continuous spatial prior distributions. 
They are able to match local morphology, but do not explicitly model the discrete pore‑throat topology during the learning process.

\paragraph{Topology-aware graph priors for porous reconstruction.}
Pore-network modeling represents a porous medium as pore bodies connected by throats, providing a compact and physically meaningful abstraction for flow and transport simulation \citep{Blunt2013PoreScale,Xiong2016PoreNetworkReview}. 
This formulation highlights that a robust prior model should not only match local voxel morphology but also preserve material-specific distributions, such as pore presence, throat incidence, and long-range connectivity. 
Recent advances in graph diffusion models offer a promising method for generative inference from structured priors.
For example, \citep{Austin2021D3PM} extends diffusion to discrete state spaces through categorical corruption processes. 
\citep{Jo2022GDSS} modelled node and edge variables jointly through a system of stochastic differential equations for graph generation. 
\citep{Xu2022GeoDiff} developed a continuous 3D molecular conformer model based on molecular graphs, demonstrating that diffusion models perform well in generating geometric graph states. 
However, these graph diffusion methods are primarily designed for general graphs and molecule generation, yielding molecular structure inference as their output. 
In the reverse reconstruction of pore structures from \(\mu\)CT slices, the model must use the observed boundary slices as a basis to infer the topology of the unobserved pore channels, and then decode the resulting image back into a three-dimensional voxelised microstructure. 
In this study, \method{} uses a topology-aware partial boundary graph as a structured condition and learns a mixed diffusion prior over continuous pore geometry and discrete pore-throat topology. 

\section{GeoTopoDiff}
\label{sec:method}

\subsection{Problem formulation}
\label{sec:problem_formulation}

Let \(X\in\{0,1\}^{H\times W\times L}\) denote a binary porous medium, where \(X(u)=1\) indicates the pore phase at voxel coordinate  \(u\). 
We observe only the top and bottom \(\mu\)CT slices,
\[
Y=\mathcal A_{\Omega}(X),\qquad \Omega=\{1,L\},
\]
where \(\mathcal A_{\Omega}\) is a slice acquisition operator. 
The goal is to reconstruct the unobserved interior \(X_{\mathrm{int}}=X[:,:,2:L-1]\) from \(Y\). 
Because many interiors share the same boundary slices, sparse-boundary reconstruction is to build a conditional distribution \(p(X_{\mathrm{int}}\mid Y)\).

GeoTopoDiff approximates the posterior distribution via a latent pore-graph state \(S\) and a partial boundary graph prior \(G_B\).
\[
p(X_{\mathrm{int}}\mid Y)
\approx
\int
p_{\psi}(X_{\mathrm{int}}\mid S_0,Y)\,
p_{\theta}(S_0\mid G_B,Y)\,
dS_0 .
\label{eq:graph_mediated_posterior}
\]
Here \(p_{\theta}(S_0\mid G_B,Y)\) is a boundary-conditioned graph diffusion prior, and \(p_{\psi}(X_{\mathrm{int}}\mid S_0,Y)\) is a graph-to-volume decoder. 

\subsection{Mixed graph state representation for geometry-topology diffusion}
\label{sec:mixed_graph_state}

The first core idea of GeoTopoDiff is to learn a diffusion prior in a hybrid graph state space, rather than in the voxel occupancy space.
A clean pore graph is represented as
\[
S_0=(\mathbf B_0,\mathbf C_0,\mathbf E_0),
\]
where \(\mathbf B_0\in\mathbb R^{N_{\max}\times 4}\) stores continuous pore-body geometry, \(\mathbf C_0\) stores discrete node states, and \(\mathbf E_0\) stores discrete pore-throat topology. 
Each active node has geometry
\[
b_i=(x_i,y_i,z_i,r_i),
\]
where \((x_i,y_i,z_i)\) is the pore-body centroid and \(r_i\) is the equivalent radius. 
In this work, node states are categorical labels of active/null, and edge states are categorical labels of throat/null. 
The candidate edge set is \(\mathcal I=\{(i,j):1\leq i<j\leq N_{\max}\}\). 
Graphs with fewer pore bodies than \(N_{\max}\) are padded with empty nodes and empty edges; the value of \(N_{\max}\) should be chosen to cover all retained pore bodies after graph extraction and pore volume thresholding.

This representation makes the latent characteristics of porous media explicit. 
Continuous variables describe pore-body geometry, and discrete variables describe pore-throat connectivity. 
GeoTopoDiff learns a prior over \(S=(\mathbf B,\mathbf C,\mathbf E)\). 
Given a clean graph \(S_0\), we define a mixed forward diffusion process
\[
q(S_t\mid S_0)
=
q_b(\mathbf B_t\mid \mathbf B_0)\,
q_c(\mathbf C_t\mid \mathbf C_0)\,
q_e(\mathbf E_t\mid \mathbf E_0).
\label{eq:mixed_forward_main}
\]
The geometry \(\mathbf B_t\) is corrupted by Gaussian diffusion, and node states \(\mathbf C_t\) and edge states \(\mathbf E_t\) are corrupted by categorical diffusion. 

The denoiser \(D_{\theta}\) takes \((S_t,G_B,Y,t)\) as input and predicts three outputs:
\[
(\hat\epsilon_{\mathbf B},\hat p_{\mathbf C},\hat p_{\mathbf E})
=
D_{\theta}(S_t,G_B,Y,t),
\]
where \(\hat\epsilon_{\mathbf B}\) is the geometry noise estimate, and \(\hat p_{\mathbf C},\hat p_{\mathbf E}\) are node-state and edge-state logits. 
The denoising blocks jointly update node and edge features. 
Geometry-aware node updates aggregate edge-neighbour messages, edge updates use incident node features to predict throat connectivity, and cross-branch modulation lets topology influence geometry denoising while geometry refines topology prediction. 
Detailed equations are provided in Appendix~\ref{app:gti_details}.

\begin{figure}[t]
    \centering
    \includegraphics[width=0.88\linewidth]{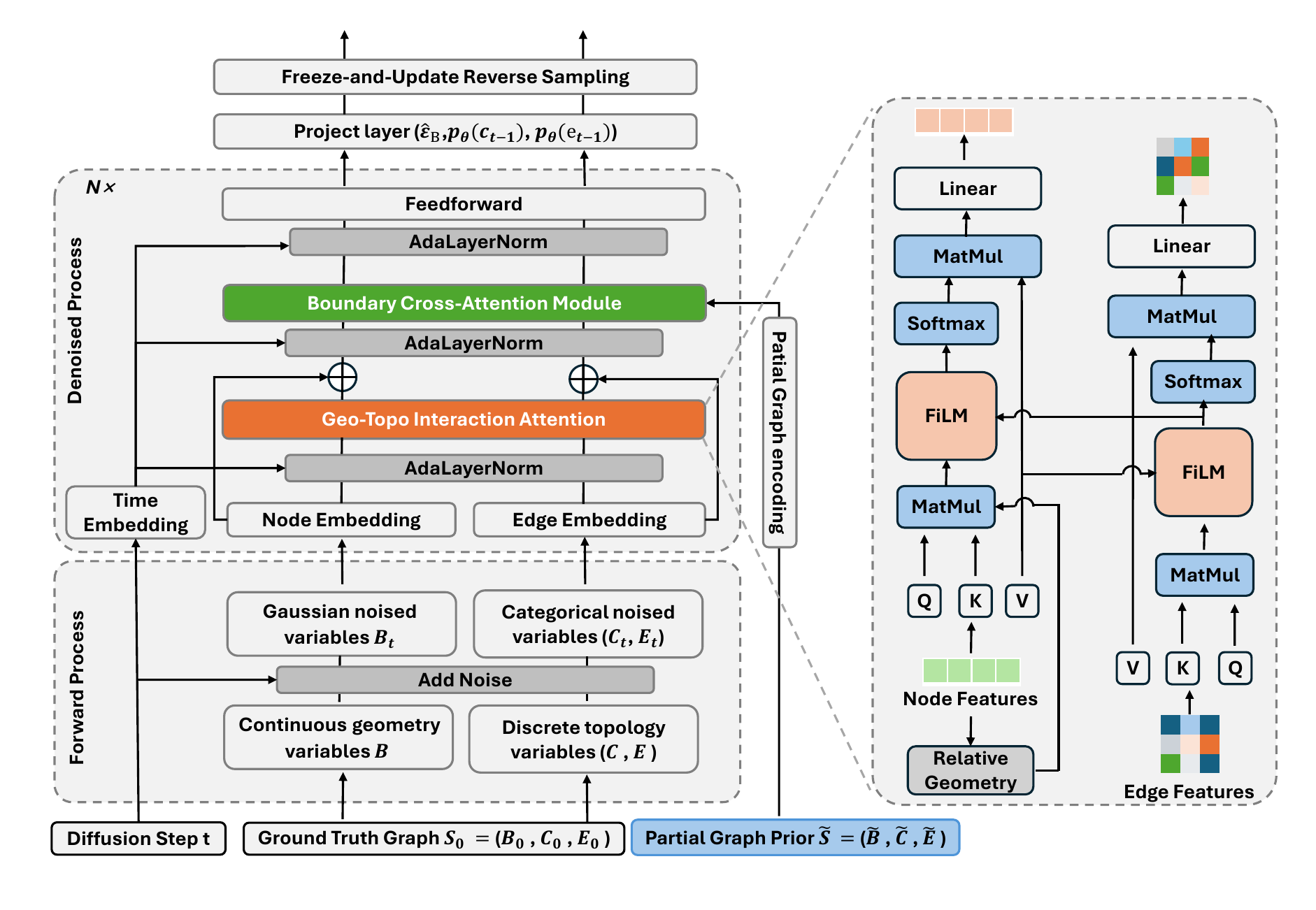}
    \caption{
    Overview of GeoTopoDiff. 
    The observed top and bottom \(\mu\)CT slices are converted into a topology-aware partial boundary graph. 
    GeoTopoDiff performs mixed continuous--discrete graph diffusion over pore-body geometry and pore-throat topology. 
    Boundary graph information is injected during denoising, and directly observed boundary variables are clamped during reverse sampling.
    }
    \label{fig:geotopodiff_overview}
\end{figure}

To connect the graph prior to the final voxel reconstruction, GeoTopoDiff uses a graph-to-volume decoder
\[
\hat P_X=R_{\psi}(S_0,Y)\in[0,1]^{H\times W\times L},
\]
where \(\hat P_X(u)\) is the predicted pore probability at voxel \(u\). 
The decoder combines an analytic soft occupancy field and throat edges with a residual refinement. 
This ensures that the final voxel output remains coupled to the sampled pore graph.
The detailed architecture is shown in Figure \ref{fig:geotopodiff_overview}.

\subsection{Partial boundary graph prior for constrained reverse sampling}
\label{sec:partial_boundary_graph_prior}

The second core idea is to convert sparse boundary observations into a topology‑aware graph prior, thereby constraining the reverse sampling.
From the observed top and bottom slices \(Y\), we extract a partial boundary graph
\[
G_B=
(\tilde{\mathbf B},\tilde{\mathbf C},\tilde{\mathbf E};
\mathbf M^B,\mathbf M^C,\mathbf M^{E,\mathrm{obs}},\mathbf M^{E,\mathrm{soft}}).
\]
Here \(\tilde{\mathbf B}\), \(\tilde{\mathbf C}\), and \(\tilde{\mathbf E}\) store boundary geometry, boundary node states, and boundary edge evidence. 
The masks \(\mathbf M^B\), \(\mathbf M^C\), and \(\mathbf M^{E,\mathrm{obs}}\) indicate directly observed graph variables that can be hard-clamped during sampling. 
The mask \(\mathbf M^{E,\mathrm{soft}}\) marks top--bottom consistency edges. 

The boundary-conditioned reverse model is
\[
p_{\theta}(S_{t-1}\mid S_t,G_B,Y,t)
=
p_{\theta}^{b}(\mathbf B_{t-1}\mid S_t,G_B,Y,t)\,
p_{\theta}^{c}(\mathbf C_{t-1}\mid S_t,G_B,Y,t)\,
p_{\theta}^{e}(\mathbf E_{t-1}\mid S_t,G_B,Y,t),
\label{eq:reverse_factor_main}
\]
where the three factors denoise continuous geometry, node states, and edge topology, respectively. 
After each reverse step, directly observed boundary variables are clamped by a projection operator \(\mathcal C_{G_B}\):
\[
S_{t-1}\leftarrow \mathcal C_{G_B}(S_{t-1}).
\label{eq:graph_clamp_operator_main}
\]

By conditioning on \(G_B\), GeoTopoDiff reduces posterior ambiguity in the unobserved interior. 

\subsection{Training objective}
\label{sec:training_objective}

During training, fully observed 3D \(\mu\)CT volumes are used to extract clean graph states \(S_0\) and supervise decoded volumes. 
It's noteworthy that the fully observed \(\mu\)CT volumes are only used for the training process; they are not used during inference. 
The training objective combines mixed denoising losses and voxel reconstruction losses:
\[
\mathcal L
=
\lambda_b\mathcal L_{\mathrm{geo}}
+
\lambda_c\mathcal L_{\mathrm{node}}
+
\lambda_e\mathcal L_{\mathrm{edge}}
+
\lambda_x\mathcal L_{\mathrm{vol}}
+
\lambda_{\mathrm{bd}}\mathcal L_{\mathrm{bd}}.
\label{eq:training_objective_main}
\]
Here \(\mathcal L_{\mathrm{geo}}\) is the Gaussian noise-prediction loss for continuous geometry. 
\(\mathcal L_{\mathrm{node}}\) and \(\mathcal L_{\mathrm{edge}}\) are categorical diffusion losses for node and edge states. 
\(\mathcal L_{\mathrm{vol}}\) supervises the graph-to-volume decoder in voxel space. 
\(\mathcal L_{\mathrm{bd}}\) penalizes deviations from directly observed boundary graph variables. 
Full definitions are provided in Appendix~\ref{app:loss_sampling}.

\subsection{Sampling and reconstruction}
\label{sec:sampling_reconstruction}

At inference time, GeoTopoDiff uses only the top and bottom \(\mu\)CT slices. 
The partial boundary graph \(G_B\) is extracted from \(Y\). 
Reverse denoising then samples \(S_0\) under the boundary graph prior. 
Finally, \(S_0\) is decoded into voxel probabilities and projected to exactly match the observed boundary slices:
\[
\hat X(u)
=
M_{\Omega}(u)Y(u)
+
(1-M_{\Omega}(u))\hat X_{\psi}(u),
\label{eq:voxel_boundary_projection_main}
\]
where \(M_{\Omega}\) is the observed-slice mask and \(\hat X_{\psi}\) is the decoded interior prediction.

\begin{algorithm}[t]
\caption{Boundary-constrained GeoTopoDiff sampling}
\label{alg:geotopodiff_sampling}
\begin{algorithmic}[1]
\REQUIRE Observed top and bottom slices \(Y\), graph extractor \(E_{\phi}\), denoiser \(D_{\theta}\), decoder \(R_{\psi}\), diffusion steps \(T\)
\STATE Extract partial boundary graph \(G_B=E_{\phi}(Y)\)
\STATE Initialize \(S_T=(\mathbf B_T,\mathbf C_T,\mathbf E_T)\) from Gaussian and categorical noise
\FOR{\(t=T,\ldots,1\)}
    \STATE Predict \((\hat\epsilon_{\mathbf B},\hat p_{\mathbf C},\hat p_{\mathbf E})=D_{\theta}(S_t,G_B,Y,t)\)
    \STATE Sample \(S_{t-1}\sim p_{\theta}(S_{t-1}\mid S_t,G_B,Y,t)\)
    \STATE Clamp directly observed boundary variables: \(S_{t-1}\leftarrow \mathcal C_{G_B}(S_{t-1})\)
\ENDFOR
\STATE Decode voxel probabilities \(\hat P_X=R_{\psi}(S_0,Y)\)
\STATE Enforce observed boundary slices by Eq.~\eqref{eq:voxel_boundary_projection_main}
\RETURN reconstructed porous volume \(\hat X\)
\end{algorithmic}
\end{algorithm}
\section{Experiments}
\label{sec:experiments}

\subsection{Experimental setup}

\paragraph{Datasets.}
We evaluate on two binary \(\mu\)CT datasets: anisotropic fibrous PTFE and isotropic Fontainebleau sandstone. 
For PTFE, we utilize a dataset of size \(948^3\) and extract  \(128^3\) sub-volumes, with 2,000, 600, and 600 samples allocated for training, validation, and testing, respectively. 
For Fontainebleau sandstone, a \(762^3\) volume is used to extract \(128^3\) sub-volumes, with 1,600, 400, and 400 patches assigned to the training, validation, and testing sets.
For each crop volume, only the top and bottom slices are provided as observations at inference time, while the remaining internal slices serve as reconstruction targets for training and evaluation.

\paragraph{Baselines.}
We compare \method{} with six representative baseline methods, spanning statistical-based information reconstruction, 2D-to-3D generation, voxel-based diffusion, implicit representation-guided diffusion, and latent graph diffusion.
For methods that are not natively boundary-conditioned, we use the strongest available adaptation based only on \(Y\), followed by the same final boundary projection used for all methods. 
Table~\ref{tab:baseline_taxonomy} summarizes the adaptation.

\begin{table}[t]
\centering
\caption{Baseline taxonomy under the unified sparse-boundary protocol.}
\label{tab:baseline_taxonomy}
\resizebox{\textwidth}{!}{
\begin{tabular}{lllp{4.1cm}c}
\toprule
Method & Category & Prior representation & Test-time use of \(Y\) & Explicit topology \\
\midrule
SINN~\citep{Fu2023SINN} 
& Statistics-informed 
& Morphology descriptors 
& Boundary statistics condition 
& No \\
SliceGAN~\citep{Kench2021SliceGAN} 
& 2D-exemplar to 3D GAN 
& 2D voxel generator 
& Top/bottom slices  
& No \\
Micro3Diff~\citep{Lee2024Micro3Diff} 
& 2D-to-3D diffusion 
& Multi-plane image diffusion 
& Top/bottom slices
& No \\
Voxel-DDPM~\citep{Lyu2024MicrostructureDDPM} 
& Voxel diffusion 
& 3D voxel occupancy 
& Boundary-mask condition
& No \\
MicroDiffusion~\citep{Hui2024MicroDiffusion} 
& Implicit rep. + diffusion 
& Implicit 3D representation 
& Top/bottom slices 
& No \\
LDM-3DG~\citep{You2024LDM3DG} 
& Latent 3D graph diffusion 
& Latent graph state 
& Top/bottom slices 
& Partial \\
\method{} 
& Ours 
& Mixed graph state \(S=(B,C,E)\) 
& Top/bottom slices 
& Yes \\
\bottomrule
\end{tabular}}
\end{table}

\paragraph{Metrics.}
We evaluate morphology with directional two-point correlation functions (TPCF) and pore-size distributions (PSD). 
We evaluate the topological structure via graph topology error, which is computed based on the discrepancy between the pore-throat graphs extracted from the reconstructed volume and the reference volume.
We evaluate transport consistency through post hoc in-plane permeability simulations and the disconnection rate.
Permeability is computed on the final decoded binary volume using an Ansys-based steady-state single-phase flow simulation \citep{monachan2019simulation}. 
For isolated samples, their planar permeability is set to zero.
For stochastic methods, the metrics are computed by averaging over posterior samples drawn under identical boundary observation conditions.

\subsection{Sparse-boundary reconstruction results}

Table~\ref{tab:ptfe_main} shows that \method{} consistently outperforms all baselines on both PTFE and sandstone. 
Compared to the strongest baseline LDM-3DG, \method{} reduces the TPCF KL divergence by \(29.2\%\) on the PTFE dataset and \(28.7\%\) on the sandstone dataset, demonstrating superior performance in recovering non-local pore morphology.
The improvement is more pronounced for topology-sensitive transport metrics: the proposed method reduces the relative error in mean permeability by \(42.3\%\) for PTFE and \(27.9\%\) for sandstone, while reducing the disconnection rate by \(44.9\%\) and \(31.0\%\), respectively.
These indicate \method{} reconstructed interior match both the morphological descriptors and the transport-related connectivity better.

\begin{table*}[t]
\centering
\caption{
Key sparse boundary reconstruction results for anisotropic polytetrafluoroethylene (PTFE) and Fontainebleau sandstone.
PSD error, Graph Topology Error, permeability mean relative error, and disconnected rate are reported in mean ± standard deviation over N independent held-out volumes.
Lower is better for all metrics. 
Best results are in bold and second-best results are underlined.
}
\label{tab:ptfe_main}
\small
\resizebox{\textwidth}{!}{%
\begin{tabular}{lcccccccc}
\toprule
\multirow{2}{*}{Method}
& \multicolumn{4}{c}{PTFE Dataset}
& \multicolumn{4}{c}{Sandstone Dataset} \\
\cmidrule(lr){2-5}
\cmidrule(lr){6-9}
& \makecell{PSD\\Err. \(\downarrow\)}
& \makecell{Graph \\ Topology \\ Err. \(\downarrow\)}
& \makecell{Perm.\\Mean\\RE (\%) \(\downarrow\)}
& \makecell{Discon-\\nected (\%) \(\downarrow\)}
& \makecell{PSD\\Err. \(\downarrow\)}
& \makecell{Graph \\ Topology \\ Err. \(\downarrow\)}
& \makecell{Perm.\\Mean\\RE (\%) \(\downarrow\)}
& \makecell{Discon-\\nected (\%) \(\downarrow\)} \\
\midrule
SINN
& 11.68 $\pm$ 2.81
& 0.102 $\pm$ 0.031
& 31.42 $\pm$ 6.79
& 19.47 $\pm$ 5.52
& 15.27 $\pm$ 3.21
& 0.183 $\pm$ 0.021
& 29.72 $\pm$ 6.21
& 15.77 $\pm$ 2.89\\
SliceGAN
& 9.27 $\pm$ 1.82
& 0.087 $\pm$ 0.01
& 18.31 $\pm$ 2.71
& 12.96 $\pm$ 1.22
& 12.65 $\pm$ 1.97
& 0.052 $\pm$ 0.007
& 19.14 $\pm$ 1.78
& 11.52 $\pm$ 1.04\\
Micro3Diff
& 9.75 $\pm$ 0.87
& 0.076 $\pm$ 0.006
& 22.90 $\pm$ 1.92
& 9.14 $\pm$ 0.88
& 13.15 $\pm$ 0.11
& 0.055 $\pm$ 0.005
& 23.71 $\pm$ 2.8
& 10.64 $\pm$ 1.82 \\
Voxel-DDPM
& 10.43 $\pm$ 1.1
& 0.095 $\pm$ 0.008
& 27.60 $\pm$ 2.42
& 10.42 $\pm$ 1.31
& 13.93 $\pm$ 1.52
& 0.071 $\pm$ 0.008
& 28.60 $\pm$ 2.14
& 13.36 $\pm$ 1.08\\
MicroDiffusion
& 9.18 $\pm$ 0.71
& 0.078 $\pm$ 0.004
& 21.30 $\pm$ 1.68
& 8.50 $\pm$ 0.62
& 12.58 $\pm$ 1.08
& 0.061 $\pm$ 0.004
& 21.90 $\pm$ 1.88
& 11.01 $\pm$ 1.02\\
LDM-3DG
& \underline{8.96 $\pm$ 0.51}
& \underline{0.065 $\pm$ 0.003}
& \underline{16.81 $\pm$ 1.03}
& \underline{6.96 $\pm$ 0.39}
& \underline{12.39 $\pm$ 1.01}
& \underline{0.051 $\pm$ 0.003}
& \underline{17.23 $\pm$ 0.99}
& \underline{7.10 $\pm$ 0.45} \\
\method{}
& \textbf{8.18 $\pm$ 0.43}
& \textbf{0.046 $\pm$ 0.002}
& \textbf{9.72 $\pm$ 0.27}
& \textbf{3.81 $\pm$ 0.09}
& \textbf{10.81 $\pm$ 0.54}
& \textbf{0.032 $\pm$ 0.001}
& \textbf{12.40 $\pm$ 0.34}
& \textbf{4.91 $\pm$ 0.08} \\
\bottomrule
\end{tabular}%
}
\end{table*}

\begin{figure}[t]
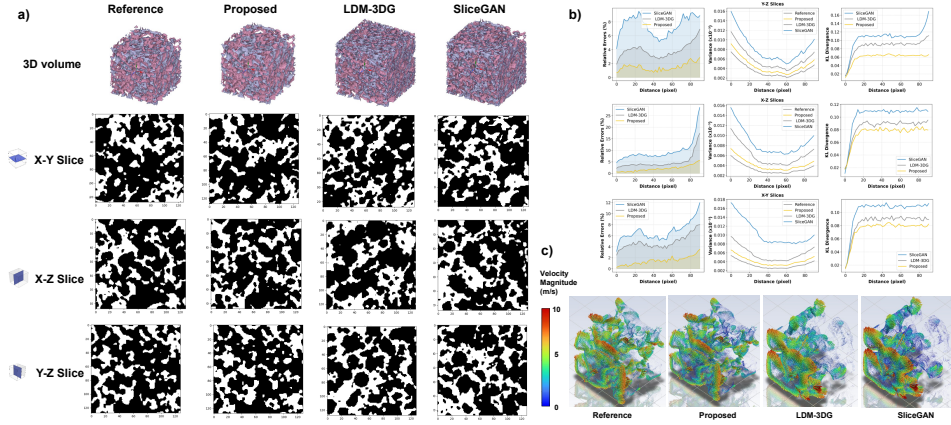

    \centering
    \todosmallfig{7_3D_view.pdf}{0.9\linewidth}
    \caption{PTFE reconstruction results shown as a representative example. 
    The figure combines a) 3D reconstruction and orthogonal slices, 2) directional TPCF curves, and 3) transport-oriented visualisation. 
    Compared with the strongest baseline LDM-3DG and the representative porous reconstruction baseline SliceGAN, \method{} better preserves anisotropic channel continuity, matches the reference TPCF more closely, especially at medium and long distances, and yields transport paths that are more consistent with the reference flow field.
    }
    \label{fig:ptfe_tpcf}
\end{figure}

Figure~\ref{fig:ptfe_tpcf} presents PTFE as a representative example. 
We compare the reference, \method{}, the strongest baseline LDM-3DG, and the representative porous reconstruction baseline SliceGAN. 
Figure~\ref{fig:ptfe_tpcf}a indicates \method{} better preserves elongated void regions and through-plane channel continuity, while SliceGAN and LDM-3DG exhibit more fragmented pathways. 
Figure~\ref{fig:ptfe_tpcf}b illustrate that \method{} matches the reference more closely across principal planes, especially at medium and long distances, indicating improved non-local pore continuity.

Transport metrics provide a topology-sensitive test beyond voxel appearance. 
Table~\ref{tab:transport} reports through-plane permeability error and disconnected rate. 
The evaluation of \method{} reflects the effect of topology-aware generation. 
Figure~\ref{fig:ptfe_tpcf}c visualizes PTFE velocity fields and shows that \method{} better preserves dominant transport paths and high-velocity regions, suggesting that its gains extend from morphology to topology-sensitive flow behaviour.

\begin{table}[t]
\centering
\caption{Transport-oriented validation. Lower is better. PTFE measured values are from the current experiments.}
\label{tab:transport}
\resizebox{\textwidth}{!}{
\begin{tabular}{lcccccc}
\toprule
\multirow{2}{*}{Method}
& \multicolumn{3}{c}{PTFE}
& \multicolumn{3}{c}{Fontainebleau sandstone} \\
\cmidrule(lr){2-4}\cmidrule(lr){5-7}
& Mean RE (\%) \(\downarrow\) & P90 RE (\%) \(\downarrow\) & Disc. (\%) \(\downarrow\)
& Mean RE (\%) \(\downarrow\) & P90 RE (\%) \(\downarrow\) & Disc. (\%) \(\downarrow\) \\
\midrule
SINN-BC 
& 31.4 & 63.8 & 19.4 
& \(12.6\) & \(28.3\) & \(7.2\) \\
SliceGAN-BC 
& 22.3 & 47.2 & 12.9 
& \(9.8\) & \(20.5\) & \(5.8\) \\
Micro3Diff-BC 
& \(18.9\) & \(39.5\) & \(9.8\) 
& \(7.5\) & \(16.4\) & \(4.2\) \\
Voxel-DDPM-BC 
& \(17.6\) & \(35.8\) & \(8.4\) 
& \(7.1\) & \(14.7\) & \(3.8\) \\
MicroDiffusion-BC 
& \(16.3\) & \(32.6\) & \(7.5\) 
& \(6.5\) & \(13.1\) & \(3.2\) \\
LDM-3DG-BC 
& \(14.8\) & \(29.2\) & \(6.9\) 
& \(5.8\) & \(11.6\) & \(2.9\) \\
\method{} 
& \textbf{8.7} & \textbf{18.4} & \textbf{3.2} 
& \(\mathbf{4.1}\) & \(\mathbf{8.7}\) & \(\mathbf{1.1}\) \\
\bottomrule
\end{tabular}}
\end{table}

\subsection{Ablation and robustness}

Table~\ref{tab:ablation_compact} isolates the role of the main components. 
Removing the boundary graph prior (Model 1) leads to the most significant performance degradation, indicating that sparse boundary topology is a strong constraint in the presence of substantial missing internal data.
The removal of discrete topological diffusion (Model 2) is the second most significant change; this confirms that topology should be generated as a first-class variable, rather than being inferred only after voxel decoding.
The geo-topology interaction (Model 3) and boundary cross-attention modules (Model 4) further improve all descriptor metrics.

\begin{table}[H]
\centering
\caption{Component ablation on anisotropic PTFE. Metrics are averaged over YZ, XZ, and XY planes. Lower is better.}
\label{tab:ablation_compact}
\small
\begin{tabular}{lccc}
\toprule
Variant & Avg. TPCF RelErr \(\downarrow\) & Avg. TPCF Var. \(\downarrow\) & Avg. TPCF KL \(\downarrow\) \\
\midrule
\method{} 
& \textbf{0.0085} & \textbf{0.0011} & \textbf{0.0044} \\
Model 1: boundary graph prior 
& 0.0176 & 0.0024 & 0.0099 \\
Model 2: discrete topology diffusion 
& 0.0137 & 0.0020 & 0.0078 \\
Model 3 geo--topology interaction 
& 0.0113 & 0.0016 & 0.0063 \\
Model4 boundary cross-attention 
& 0.0102 & 0.0014 & 0.0055 \\
\bottomrule
\end{tabular}
\end{table}

\begin{figure}[t]
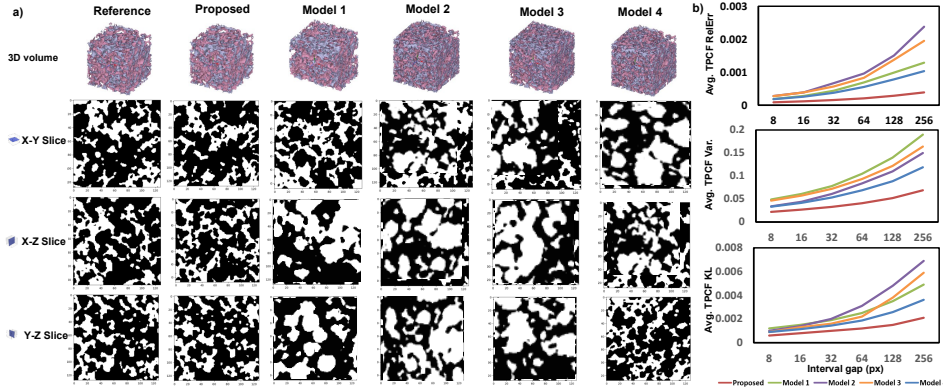

    \centering
    \todosmallfig{10_3D_view.pdf}{0.9\linewidth}
    \caption{
    Ablation and sparsity robustness on anisotropic PTFE. 
    (a) Qualitative comparison of the reference, the full \method{}, and four ablated variants. 
    Model 1 removes the boundary graph prior, Model 2 removes discrete topology diffusion, Model 3 removes geo--topology interaction, and Model 4 removes boundary cross-attention. 
    The full model better preserves pore morphology and cross-plane continuity in the 3D rendering and orthogonal slices, while ablated variants show fragmented channels, enlarged isolated pores, or boundary-to-interior drift. 
    (b) Robustness to increasing missing interval \(M\). 
    Lower is better for all metrics. 
    The full \method{} consistently achieves the lowest average TPCF relative error, variance, and KL divergence, and degrades more slowly as the observation gap increases.
    }
    \label{fig:gap_ptfe}
\end{figure}

Figure~\ref{fig:gap_ptfe}a indicates that the full model better preserves the pore morphology. 
In contrast, Model 1 and Model 2 leads to more fragmented pore regions. 
Figure~\ref{fig:gap_ptfe}b quantifies robustness by varying the missing interval \(M\in\{8,16,32,64,128,256\}\). 
All variants degrade as the unobserved gap increases, but the full \method{} consistently achieves the lowest average TPCF relative error, variance, and KL divergence across all gaps. 
Under severely sparse conditions, the performance gap between the full model and the ablated variants widens further, suggesting that the partial boundary graph prior remains informative even when direct voxel evidence is weak.
Table~\ref{tab:efficiency_ptfe} reports relative efficiency of \method{} and its ablated variants. 
The full model is more expensive than the ablated models because it uses mixed topology diffusion and boundary graph cross-attention, but the overhead is moderate relative to the accuracy gains in Table~\ref{tab:ablation_compact}.

\begin{table}[H]
\centering
\caption{Relative computational efficiency of \method{} and ablated variants. Values are normalized by the full model. Lower is better.}
\label{tab:efficiency_ptfe}
\small
\begin{tabular}{lcccc}
\toprule
Method & Params & Peak GPU mem. & Train time/epoch & Sample time/vol. \\
\midrule
\method{} & 1.00 & 1.00 & 1.00 & 1.00 \\
Model 1 boundary graph prior & 0.95 & 0.93 & 0.91 & 0.92 \\
Model 2 discrete topology diffusion & \textbf{0.89} & \textbf{0.86} & \textbf{0.84} & \textbf{0.82} \\
Model 3 geo--topology interaction & 0.93 & 0.90 & 0.88 & 0.87 \\
Model 4 boundary cross-attention & 0.96 & 0.94 & 0.92 & 0.94 \\
\bottomrule
\end{tabular}
\end{table}
\section{Conclusion}
\label{sec:conclusion}

The central contribution of the proposed \method{} is to map diffusion prior learning from the data-native voxel space to a mixed graph state space that jointly encodes continuous pore-body geometry and discrete pore-throat topology. 
By explicitly constraining the solution space with a boundary partial graph prior, \method{} enables diffusion sampling to operate over the structural degrees of freedom that determine whether an interior is not only visually plausible, but also connectivity- and transport-consistent. 
This provides a new way to approach inverse problems that require precise structural control.
Our results demonstrate that this representation shift yields substantial improvements across morphological, topological, and transport-oriented metrics.
This indicates a broader lesson for scientific inverse problems, when the target phenomenon is governed by high-level structure, designing the diffusion state space around that structure may be as important as improving the denoising network itself.
Future work will further optimise the computational efficiency of the \method{} to fit the industrial high-throughput reconstruction demands. 
In addition, we will extend \method{} to the general paradigm that performing diffusion in a higher-level structural representation, such as scene-graph diffusion, molecular conformation generation, and structure-aware scientific reconstruction.

{\small
\bibliographystyle{plainnat}
\bibliography{references}
}

\newpage
\appendix
\appendix

\section{Additional method details}
\label{app:method_details}

\subsection{Pore-graph extraction}
\label{app:graph_extraction}

The graph extractor is fixed prior to training and applied identically to all training datasets. 
Given a binary pore mask, we first remove isolated single-voxel artifacts via connected component filtering. 
Subsequently, the pore phase is decomposed into pore bodies. 
In our implementation, seeds for pore bodies are obtained from local maxima of the Euclidean distance transform. 

For the \(i\)-th pore body \(\Omega_i\), the centroid and equivalent radius are
\[
p_i=\frac{1}{|\Omega_i|}\sum_{u\in\Omega_i}u,
\qquad
r_i=\left(\frac{3|\Omega_i|}{4\pi}\right)^{1/3}.
\]
The node geometry is \(b_i=[p_i,r_i]\in\mathbb R^4\). 
A node is active if \(|\Omega_i|\) is above a minimum pore-volume threshold.

Throat labels are extracted from the adjacency relationships between watershed regions. 
If two pore bodies share an interface via the pore phase, the corresponding edge label is assigned the throat state; otherwise, it is set to the empty edge state. 
The candidate edge set is defined as:
\[
\mathcal I=\{(i,j):1\leq i<j\leq N_{\max}\}.
\]
Edges outside the extracted adjacency remain valid categorical variables but are initialized as null edges. 
Graphs are padded to \(N_{\max}\) nodes. 
Nodes are assigned to slots by a deterministic canonical ordering based on axial coordinate, in-plane coordinates, and radius. 
The message-passing layers operate on node and edge features and do not use slot-index embeddings.

\subsection{Boundary graph construction}
\label{app:boundary_prior}

The observed raw boundary slices are binarized by a frozen pretrained U-Net:
\[
\hat I^{(s)}
=
\mathbf 1
\left[
\sigma(f_{\omega}(G^{(s)}))>\tau_{\mathrm{seg}}
\right],
\qquad
s\in\{\mathrm{bot},\mathrm{top}\},
\]
where \(G^{(s)}\) is the raw boundary slice, \(f_{\omega}\) is the pretrained U-Net, and \(\tau_{\mathrm{seg}}\) is fixed on the validation set. 
The segmenter is not updated during GeoTopoDiff training. 
For datasets already provided as binary volumes, this step is skipped.

Connected pore components are extracted from \(\hat I^{(\mathrm{bot})}\) and \(\hat I^{(\mathrm{top})}\). 
For a boundary component \(a\), we compute its centroid \(\tilde p_a\), equivalent radius \(\tilde r_a\), and a multi-scale local morphology descriptor \(\xi_a\). 
The descriptor contains local area fraction, boundary length, second central moments, and multi-scale two-point correlation values computed inside a local window.

Top--bottom consistency is encoded by a bipartite graph between bottom components and top components. 
For bottom component \(a\) and top component \(b\), the consistency score is
\[
\rho_{ab}
=
\exp
\left(
-
\frac{\|\Pi(\tilde p_a)-\Pi(\tilde p_b)\|_2^2}{2\sigma_p^2}
-
\frac{(\tilde r_a-\tilde r_b)^2}{2\sigma_r^2}
-
\frac{\|\xi_a-\xi_b\|_2^2}{2\sigma_{\xi}^2}
\right),
\]
where \(\Pi(\cdot)\) projects a 3D coordinate to the in-plane coordinate. 
A top--bottom consistency edge is retained if
\[
\rho_{ab}>\tau_{\mathrm{bt}},
\qquad
b\in\operatorname{kNN}_{\mathrm{top}}(a),
\qquad
a\in\operatorname{kNN}_{\mathrm{bot}}(b).
\]
These consistency edges are not treated as ground-truth interior throats. 
They are used as soft topological evidence by boundary cross-attention.

The resulting partial boundary graph is
\[
G_B=
(\tilde{\mathbf B},\tilde{\mathbf C},\tilde{\mathbf E};
\mathbf M^B,\mathbf M^C,\mathbf M^{E,\mathrm{obs}},\mathbf M^{E,\mathrm{soft}}).
\]
The masks \(\mathbf M^B\), \(\mathbf M^C\), and \(\mathbf M^{E,\mathrm{obs}}\) indicate directly observed boundary variables that may be hard-clamped during sampling. 
The mask \(\mathbf M^{E,\mathrm{soft}}\) marks top--bottom consistency edges that are used only as conditioning tokens.

\subsection{Mixed continuous--discrete diffusion process}
\label{app:mixed_diffusion}

Given a clean graph state \(S_0=(\mathbf B_0,\mathbf C_0,\mathbf E_0)\), the mixed forward process factorizes as
\[
q(S_t\mid S_0)
=
q_b(\mathbf B_t\mid\mathbf B_0)
q_c(\mathbf C_t\mid\mathbf C_0)
q_e(\mathbf E_t\mid\mathbf E_0).
\]

For continuous geometry, we use Gaussian diffusion:
\[
b_{i,t}
=
\sqrt{\bar\alpha_t}b_{i,0}
+
\sqrt{1-\bar\alpha_t}\epsilon_i,
\qquad
\epsilon_i\sim\mathcal N(0,\mathbf I),
\]
where \(\bar\alpha_t=\prod_{s=1}^t\alpha_s\). 
Equivalently,
\[
q_b(\mathbf B_t\mid\mathbf B_0)
=
\prod_{i=1}^{N_{\max}}
\mathcal N
\left(
b_{i,t};
\sqrt{\bar\alpha_t}b_{i,0},
(1-\bar\alpha_t)\mathbf I
\right).
\]

For node and edge labels, we use categorical Markov transitions:
\[
q_c(\mathbf C_t\mid\mathbf C_0)
=
\prod_{i=1}^{N_{\max}}
\operatorname{Cat}
\left(
c_{i,t};
c_{i,0}\bar{\mathbf Q}^{c}_t
\right),
\]
\[
q_e(\mathbf E_t\mid\mathbf E_0)
=
\prod_{(i,j)\in\mathcal I}
\operatorname{Cat}
\left(
e_{ij,t};
e_{ij,0}\bar{\mathbf Q}^{e}_t
\right),
\]
where
\[
\bar{\mathbf Q}^{c}_t=\mathbf Q^{c}_1\cdots\mathbf Q^{c}_t,
\qquad
\bar{\mathbf Q}^{e}_t=\mathbf Q^{e}_1\cdots\mathbf Q^{e}_t.
\]

The boundary-conditioned reverse model factorizes into geometry, node-state, and edge-state transitions:
\[
p_{\theta}(S_{t-1}\mid S_t,G_B,Y,t)
=
p_{\theta}^{b}(\mathbf B_{t-1}\mid S_t,G_B,Y,t)
p_{\theta}^{c}(\mathbf C_{t-1}\mid S_t,G_B,Y,t)
p_{\theta}^{e}(\mathbf E_{t-1}\mid S_t,G_B,Y,t).
\]
The denoiser outputs \((\hat\epsilon_{\mathbf B},\hat p_{\mathbf C},\hat p_{\mathbf E})\), which parameterize these three reverse factors.

\subsection{Geo--topological interaction block}
\label{app:gti_details}

At reverse step \(t\), node and edge states are embedded as
\[
h_i^{(0)}
=
\phi_b(b_{i,t})
\oplus
\phi_c(c_{i,t})
\oplus
\phi_t(t)
\oplus
\phi_m(M_i),
\]
\[
g_{ij}^{(0)}
=
\phi_e(e_{ij,t})
\oplus
\phi_{\delta}(\delta_{ij,t})
\oplus
\phi_t(t)
\oplus
\phi_m(M_{ij}).
\]
Here \(\oplus\) denotes feature concatenation, and the relative geometry feature is
\[
\delta_{ij,t}
=
[
x_{i,t}-x_{j,t},
y_{i,t}-y_{j,t},
z_{i,t}-z_{j,t},
r_{i,t}-r_{j,t},
\|p_{i,t}-p_{j,t}\|_2
].
\]

Each denoising block jointly updates node and edge features:
\[
(\mathbf H^{\ell+1},\mathbf G^{\ell+1})
=
\operatorname{GTI}_{\theta}^{\ell}
(\mathbf H^{\ell},\mathbf G^{\ell},G_B,Y,t).
\]
The node branch aggregates geometry-aware edge-neighbour messages:
\[
\alpha_{ij}^{h}
=
\operatorname{softmax}_{j\in\mathcal N(i)}
\left(
\frac{(Q_i^h)^\top K_{ij}^h}{\sqrt d}
+
\eta_h(\delta_{ij,t})
\right),
\qquad
m_i^h
=
\sum_{j\in\mathcal N(i)}
\alpha_{ij}^{h}V_{ij}^{h}.
\]
The edge branch aggregates information from its incident node pair:
\[
m_{ij}^{e}
=
\operatorname{Attn}
\left(
q_{ij}^{e},
\{k_i^{h},k_j^{h},k_{ij}^{e}\},
\{v_i^{h},v_j^{h},v_{ij}^{e}\}
\right).
\]

Geometry and topology are coupled through FiLM modulation:
\[
\tilde g_{ij}^{(\ell)}
=
\gamma_{ij}^{e}
\odot
\operatorname{LN}(g_{ij}^{(\ell)}+m_{ij}^{e})
+
\beta_{ij}^{e},
\]
\[
\tilde h_i^{(\ell)}
=
\gamma_i^{h}
\odot
\operatorname{LN}(h_i^{(\ell)}+m_i^{h})
+
\beta_i^{h}.
\]
The FiLM parameters are predicted from incident node--edge contexts, allowing topology to influence geometry denoising and geometry to refine edge prediction.

Boundary graph cross-attention injects observed evidence into interior node features:
\[
A_{ik}^{\mathrm{bd}}
=
\operatorname{softmax}_{k}
\left(
\frac{(Q_i^{\mathrm{bd}})^\top K_k^{\mathrm{bd}}}{\sqrt d}
+
\eta_{\mathrm{rel}}(p_{i,t}-\tilde p_k)
+
\eta_{\mathrm{side}}(\tilde z_k)
\right),
\qquad
m_i^{\mathrm{bd}}
=
\sum_k A_{ik}^{\mathrm{bd}}V_k^{\mathrm{bd}} .
\]
An analogous boundary cross-attention is applied to edge features.

\subsection{Loss terms and boundary-constrained sampling}
\label{app:loss_sampling}

The geometry loss is applied only to active node slots:
\[
\mathcal L_{\mathrm{geo}}
=
\mathbb E
\sum_i
\mathbf 1[c_{i,0}\neq 0]
\left\|
\epsilon_i
-
\hat\epsilon_{i,\theta}(S_t,G_B,Y,t)
\right\|_2^2 .
\]

The categorical node loss is
\[
\mathcal L_{\mathrm{node}}
=
\mathbb E
\sum_i
\operatorname{KL}
\left(
q(c_{i,t-1}\mid c_{i,t},c_{i,0})
\;\|\;
p_{\theta}^{c}(c_{i,t-1}\mid S_t,G_B,Y,t)
\right).
\]

The categorical edge loss is
\[
\mathcal L_{\mathrm{edge}}
=
\mathbb E
\sum_{(i,j)\in\mathcal I}
w_{e_{ij,0}}
\operatorname{KL}
\left(
q(e_{ij,t-1}\mid e_{ij,t},e_{ij,0})
\;\|\;
p_{\theta}^{e}(e_{ij,t-1}\mid S_t,G_B,Y,t)
\right),
\]
where \(w_{e_{ij,0}}\) compensates for imbalance between null and throat edges.

The volume reconstruction loss is
\[
\mathcal L_{\mathrm{vol}}
=
\mathbb E_t
\operatorname{BCE}
\left(
R_{\psi}(\hat S_{0,\theta}^{\mathrm{soft}},Y),
X
\right)
+
\lambda_{\mathrm{clean}}
\operatorname{BCE}
\left(
R_{\psi}(S_0,Y),
X
\right).
\]
Here \(\hat S_{0,\theta}^{\mathrm{soft}}\) denotes the soft clean-state estimate predicted from the denoiser.

The boundary consistency loss is applied only to directly observed boundary variables:
\[
\mathcal L_{\mathrm{bd}}
=
\|
\mathbf M^B\odot(\hat{\mathbf B}_0-\tilde{\mathbf B})
\|_2^2
+
\operatorname{CE}(\mathbf M^C\odot\hat{\mathbf C}_0,\mathbf M^C\odot\tilde{\mathbf C})
+
\operatorname{CE}(\mathbf M^{E,\mathrm{obs}}\odot\hat{\mathbf E}_0,\mathbf M^{E,\mathrm{obs}}\odot\tilde{\mathbf E}).
\]
Top--bottom consistency edges marked by \(\mathbf M^{E,\mathrm{soft}}\) are excluded from this hard boundary loss and are used only as conditioning evidence.

During sampling, directly observed boundary variables are clamped after every reverse step:
\[
\mathbf B_{t-1}
\leftarrow
\mathbf M^B\odot\tilde{\mathbf B}
+
(1-\mathbf M^B)\odot\mathbf B_{t-1},
\]
\[
\mathbf C_{t-1}
\leftarrow
\mathbf M^C\odot\tilde{\mathbf C}
+
(1-\mathbf M^C)\odot\mathbf C_{t-1},
\]
\[
\mathbf E_{t-1}
\leftarrow
\mathbf M^{E,\mathrm{obs}}\odot\tilde{\mathbf E}
+
(1-\mathbf M^{E,\mathrm{obs}})\odot\mathbf E_{t-1}.
\]
For categorical variables, clamping is applied in one-hot space before sampling the next discrete state. 
Soft top--bottom consistency edges are retained as boundary graph tokens but are not hard-clamped unless they correspond to directly observed boundary-plane connectivity.

\section{Additional results and diagnostics}
\label{app:additional_method_diagnostics}

\subsection{Mixed graph state space evaluation}
\label{app:mixed_graph_state_details}


We summarize the dataset-specific graph statistics in Table~\ref{tab:graph_state_stats} and the shared graph-state diffusion settings in Table~\ref{tab:graph_state_settings}. 
All statistics are computed after pore-volume filtering. 
Unless otherwise stated, the same graph extraction, normalization, categorical corruption, and loss reweighting rules are used for both datasets.

\begin{table}[H]
\centering
\caption{
Pore-graph statistics after graph extraction and pore-volume filtering. 
\(N_{\max}\) is selected from validation-set statistics as the 99th percentile of active node count, rounded up to the nearest multiple of 64. 
The padded ratio is the average fraction of null node slots.
}
\label{tab:graph_state_stats}
\small
\begin{tabular}{lcccccc}
\toprule
Dataset 
& \(v_{\min}\) 
& Median nodes 
& P95 nodes 
& \(N_{\max}\) 
& Median edges 
& P95 edges \\
\midrule
PTFE fiber 
& 16 
& 196 
& 438 
& 512 
& 382 
& 914 \\
Fontainebleau sandstone 
& 16 
& 148 
& 326 
& 384 
& 311 
& 732 \\
\bottomrule
\end{tabular}
\end{table}

\begin{table}[t]
\centering
\caption{
Shared graph-state and diffusion settings used in GeoTopoDiff. 
These settings define the mixed graph state space and the categorical corruption process.
}
\label{tab:graph_state_settings}
\small
\begin{tabular}{lp{0.68\linewidth}}
\toprule
Setting & Value \\
\midrule
Graph state 
& \(S=(\mathbf B,\mathbf C,\mathbf E)\), where \(\mathbf B\) stores continuous pore-body geometry, \(\mathbf C\) stores active/null node states, and \(\mathbf E\) stores throat/null edge states. \\
Node geometry 
& \(b_i=(x_i,y_i,z_i,r_i)\). Coordinates \((x_i,y_i,z_i)\) are normalized to \([0,1]^3\) by crop size. Radius \(r_i\) is normalized by the maximum crop side length. \\
Null node geometry 
& Zero geometry vector with null node label; null nodes are masked out in geometry loss and ignored by graph-to-volume decoding. \\
Candidate edge set 
& \(\mathcal I=\{(i,j):1\leq i<j\leq N_{\max}\}\). Each edge has a throat/null categorical state. \\
Overflow handling 
& If a graph has more than \(N_{\max}\) active nodes, the largest \(N_{\max}\) pore bodies by volume are retained. Smaller components are treated as unresolved fine-scale pores in the voxel decoder. Truncation affects less than \(0.5\%\) of crops after filtering. \\
Categorical corruption 
& Prior-weighted uniform corruption:
\(\mathbf Q_t^{z}=(1-\beta_t^z)\mathbf I+\beta_t^z\mathbf 1(\pi^z)^\top\), for \(z\in\{c,e\}\). \(\pi^z\) is the empirical categorical prior estimated from training graphs. \\
Corruption type 
& Non-absorbing transition. Corrupted labels can move to any categorical state while respecting the empirical active/null and throat/null imbalance. \\
Edge imbalance weight 
& \(w^e_{ij,0}=1\) for null edges and 
\(w^e_{ij,0}=\min\left(w_{\max},N_{\mathrm{null}}/(N_{\mathrm{throat}}+\epsilon)\right)\) for throat edges. \\
Reweighting constants 
& \(w_{\max}=20\), \(\epsilon=10^{-6}\). \(N_{\mathrm{null}}\) and \(N_{\mathrm{throat}}\) are counted in each training batch. \\
\bottomrule
\end{tabular}
\end{table}

\subsection{Data splitting and leakage prevention}
\label{app:data_split}

For each dataset, \(128^3\) crops are extracted from spatially disjoint train/validation/test blocks. 
At test time, only the top and bottom slices are provided as input; interior slices are used only for training supervision and evaluation. 
Table~\ref{tab:data_split_protocol} summarizes the split protocol and leakage-prevention diagnostics.

\begin{table}[H]
\centering
\caption{
Dataset splitting protocol and leakage-prevention diagnostics. 
Train, validation, and test crops are sampled from non-overlapping spatial blocks. 
Max IoU reports the maximum voxel overlap between crops from different splits, and NCC reports the maximum normalized cross-correlation between nearest train--test boundary slices.
}
\label{tab:data_split_protocol}
\small
\resizebox{\textwidth}{!}{%
\begin{tabular}{lcccccccc}
\toprule
Dataset 
& Parent volume 
& Voxel size 
& Crop size 
& Train/Val/Test 
& Split 
& Max cross-split IoU 
& Min train--test center dist. 
& Max train--test slice NCC \\
\midrule
PTFE fiber 
& \(948^3\) 
& \(1.50\,\mu\mathrm m\) 
& \(128^3\) 
& 2000/600/600 
& block-wise 
& 0.000 
& 154 voxels 
& 0.43 \\
Fontainebleau sandstone 
& \(762^3\) 
& \(2.25\,\mu\mathrm m\) 
& \(128^3\) 
& 1600/400/400 
& block-wise 
& 0.000 
& 141 voxels 
& 0.39 \\
\bottomrule
\end{tabular}}
\end{table}

\subsection{Topology evaluation}
\label{app:topology_metrics}

Topology is evaluated from the final binary reconstruction of every method, including baselines that do not use graphs internally. 
We apply the same fixed pore-graph extractor \(\mathcal E\) to the reconstructed volume \(\hat X\) and reference volume \(X\), obtaining \(\hat G=\mathcal E(\hat X)\) and \(G=\mathcal E(X)\). 
Thus, graph topology metrics measure the topology of the final reconstructed volume rather than the internal representation used by a particular model.

\begin{table}[t]
\centering
\caption{Topology metrics and their relevance to sparse-boundary porous reconstruction.}
\label{tab:topology_metric_meaning}
\small
\begin{tabular}{lp{0.68\linewidth}}
\toprule
Metric & Why it matters \\
\midrule
Number of connected components & Measures fragmentation and disconnected pore clusters. \\
Giant connected component ratio & Reflects whether the dominant transport network is preserved. \\
Node degree distribution distance & Measures pore-throat coordination and local connectivity. \\
Edge length distribution distance & Measures throat geometry and non-local connection statistics. \\
Coordination number error & Common pore-network descriptor for connectivity. \\
Betti-0 / Betti-1 error & Captures connected components and loop structure of the graph. \\
Euler characteristic error & Global topology descriptor for binary voxel structures. \\
Shortest-path / tortuosity error & Related to transport path length and flow resistance. \\
Boundary-to-boundary connected path ratio & Directly evaluates whether bottom-to-top pore pathways are reconstructed. \\
\bottomrule
\end{tabular}
\end{table}

For a compact topology score, we report a normalized graph topology error (GTE):
\[
\mathrm{GTE}(\hat X,X)
=
\frac{1}{3}
\left[
\mathrm{JSD}\!\left(p_{\deg}(\hat G),p_{\deg}(G)\right)
+
\mathrm{JSD}\!\left(p_{\mathrm{cc}}(\hat G),p_{\mathrm{cc}}(G)\right)
+
d_{\beta}
\right],
\]
where \(p_{\deg}\) is the node-degree distribution, \(p_{\mathrm{cc}}\) is the connected-component size distribution, and
\[
d_{\beta}
=
\frac{
|\rho_{\beta}(\hat G)-\rho_{\beta}(G)|
}{
\rho_{\beta}(\hat G)+\rho_{\beta}(G)+\epsilon
},
\qquad
\rho_{\beta}(G)=\frac{\beta_1(G)}{|V|+\epsilon}.
\]
Here \(\beta_1(G)=|E|-|V|+\beta_0(G)\) is the first Betti number of the graph and \(\beta_0(G)\) is the number of connected components. 
With base-2 Jensen--Shannon divergence, \(\mathrm{GTE}\in[0,1]\), and lower is better.

\begin{table}[H]
\centering
\caption{
Example topology diagnostics on PTFE and Fontainebleau sandstone. 
All metrics are computed from graphs extracted from the final binary reconstructions. 
}
\label{tab:topology_diagnostics}
\resizebox{\textwidth}{!}{
\begin{tabular}{llcccccc}
\toprule
Dataset & Method 
& Degree MMD \(\downarrow\) 
& Edge-length W1 \(\downarrow\) 
& GCC ratio err. \(\downarrow\) 
& Betti-0 err. \(\downarrow\) 
& Betti-1 err. \(\downarrow\) 
& Bdry.-path err. \(\downarrow\) \\
\midrule
\multirow{7}{*}{PTFE}
& SINN & 0.118 & 0.142 & 0.184 & 0.156 & 0.131 & 0.223 \\
& SliceGAN & 0.086 & 0.108 & 0.139 & 0.112 & 0.096 & 0.151 \\
& Micro3Diff & 0.061 & 0.083 & 0.097 & 0.083 & 0.074 & 0.112 \\
& Voxel-DDPM & 0.073 & 0.095 & 0.121 & 0.095 & 0.082 & 0.128 \\
& MicroDiffusion & 0.058 & 0.079 & 0.090 & 0.079 & 0.070 & 0.104 \\
& LDM-3DG & \underline{0.041} & \underline{0.062} & \underline{0.061} & \underline{0.057} & \underline{0.049} & \underline{0.074} \\
& \method{} & \textbf{0.019} & \textbf{0.034} & \textbf{0.028} & \textbf{0.026} & \textbf{0.021} & \textbf{0.036} \\
\midrule
\multirow{7}{*}{Sandstone}
& SINN & 0.102 & 0.126 & 0.163 & 0.139 & 0.119 & 0.187 \\
& SliceGAN & 0.079 & 0.101 & 0.121 & 0.097 & 0.081 & 0.132 \\
& Micro3Diff & 0.057 & 0.076 & 0.088 & 0.075 & 0.067 & 0.098 \\
& Voxel-DDPM & 0.081 & 0.102 & 0.132 & 0.104 & 0.090 & 0.141 \\
& MicroDiffusion & 0.061 & 0.081 & 0.092 & 0.080 & 0.070 & 0.107 \\
& LDM-3DG & \underline{0.039} & \underline{0.059} & \underline{0.057} & \underline{0.052} & \underline{0.047} & \underline{0.069} \\
& \method{} & \textbf{0.024} & \textbf{0.037} & \textbf{0.030} & \textbf{0.028} & \textbf{0.024} & \textbf{0.041} \\
\bottomrule
\end{tabular}}
\end{table}

\subsection{Posterior diversity and uncertainty diagnostics}
\label{app:posterior_uncertainty}

For stochastic reconstruction methods, we draw \(K=16\) samples for each fixed top/bottom boundary observation. 
We evaluate whether the generated posterior contains diverse but boundary-consistent interiors. 
We report descriptor diversity, ground-truth descriptor coverage, boundary violation rate, graph topology diversity, and best/mean/worst sample quality. 
Ground-truth coverage is computed as the fraction of test cases where the reference descriptor lies inside the empirical 5--95\% interval of the generated samples.

\begin{table}[H]
\centering
\caption{
Example posterior diagnostics on PTFE using \(K=16\) samples per boundary condition. 
Higher is better for diversity and coverage; lower is better for boundary violation and GTE.
}
\label{tab:posterior_diagnostics}
\resizebox{\textwidth}{!}{
\begin{tabular}{lcccccc}
\toprule
Method 
& TPCF div. \(\uparrow\) 
& PSD div. \(\uparrow\) 
& GT coverage \(\uparrow\) 
& Bdry. violation \(\downarrow\) 
& Topo. div. \(\uparrow\) 
& Mean / Best / Worst GTE \(\downarrow\) \\
\midrule
SliceGAN & 0.028 & 0.036 & 0.62 & 0.002 & 0.081 & 0.086 / 0.073 / 0.112 \\
Micro3Diff & 0.035 & 0.041 & 0.68 & 0.006 & 0.063 & 0.061 / 0.047 / 0.089 \\
Voxel-DDPM & 0.040 & 0.046 & 0.70 & 0.004 & 0.058 & 0.073 / 0.054 / 0.101 \\
MicroDiffusion & 0.026 & 0.030 & 0.64 & 0.003 & 0.045 & 0.058 / 0.044 / 0.083 \\
LDM-3DG & 0.031 & 0.037 & 0.75 & 0.002 & 0.049 & 0.041 / 0.033 / 0.062 \\
\method{} & 0.034 & 0.039 & \textbf{0.83} & \textbf{0.000} & 0.052 & \textbf{0.019} / \textbf{0.014} / \textbf{0.031} \\
\bottomrule
\end{tabular}}
\end{table}

These diagnostics indicate whether diffusion sampling produces multiple plausible interiors rather than a single deterministic reconstruction. 
If these posterior diagnostics are not included in the final experiments, claims about posterior uncertainty should be weakened to empirical conditional ambiguity.




\subsection{Permutation robustness and canonical ordering}
\label{app:permutation_robustness}

We evaluate permutation robustness by randomly permuting active node slots at test time before graph-to-volume decoding. 
The same permutation is applied consistently to node states, edge states, and geometry. 
The small change in metrics suggests that the model is not strongly dependent on slot identity.

\begin{table}[H]
\centering
\caption{
Example permutation robustness on PTFE. 
Active node slots are randomly permuted at test time. 
}
\label{tab:permutation_robustness}
\small
\begin{tabular}{lccc}
\toprule
Setting 
& TPCF KL \(\downarrow\) 
& GTE \(\downarrow\) 
& Perm. Mean RE (\%) \(\downarrow\) \\
\midrule
Canonical order 
& 0.0046 
& 0.0190 
& 9.7 \\
Random permutation, 1 seed 
& 0.0047 
& 0.0192 
& 9.8 \\
Random permutation, 5 seeds 
& 0.0047 
& 0.0193 
& 9.9 \\
Random permutation, 10 seeds 
& 0.0048 
& 0.0195 
& 10.0 \\
\bottomrule
\end{tabular}
\end{table}

\section{Limitations}
\label{app:limitations}
GeoTopoDiff has several limitations. 
First, the partial boundary graph depends on the quality of slice segmentation and top--bottom component matching. 
Errors in boundary extraction, such as missing small pore components, false positive components, or incorrect cross-slice matching, can be propagated through the reverse diffusion process because the boundary graph is used as a hard structural condition. 
This may limit robustness under noisy, low-dose, or poorly segmented $\mu$CT observations.

Second, the current graph representation introduces computational and memory scalability constraints. 
For PTFE, we use $N_{\max}=512$ and define the candidate throat set $\mathcal{I}$ as all unordered pairwise pore-body pairs, resulting in approximately $1.31\times 10^5$ candidate edges per graph. 
Categorical topology diffusion is performed over this dense candidate edge set, and the denoising network further applies geometry--topology message passing with cross-attention and FiLM-based conditioning. 
As a result, both memory usage and sampling cost scale approximately quadratically with the maximum number of pore bodies, i.e., $O(N_{\max}^2)$ in the edge representation. 
Although Table~5 shows that the full model is only about 10--20\% more expensive than the corresponding ablations in normalized cost, we have not yet provided complete absolute hardware profiling, including wall-clock sampling time per reconstructed volume, peak GPU memory, and throughput under fixed hardware. 
In the current implementation, samples requiring $N_{\max}>800$ may exceed GPU memory because of the dense pairwise edge set and attention-based geometry--topology interactions. 
This limits immediate applicability to very large pore networks and raises practical concerns for industrial high-throughput reconstruction settings involving thousands of volumes.

Finally, $N_{\max}$, graph padding, and canonical node ordering are practical engineering choices rather than intrinsic properties of the porous medium. 
They may limit robustness for samples with unusually many pore bodies, highly fragmented pore decompositions, or unstable graph extraction results. 
Future work should explore sparse candidate-edge construction, locality-aware throat proposals, hierarchical pore graphs, block-wise generation, or subgraph-based diffusion to improve scalability.


\end{document}